\documentclass[preprint]{elsarticle}
\usepackage{multirow}
\usepackage{graphicx}
\usepackage{amsmath}
\usepackage{hyperref}
\usepackage{subcaption}
\usepackage[algosection]{algorithm2e}
\usepackage{amssymb}

\newdefinition{example}{Example}
\newdefinition{envirolist}{Procedure}
\journal{}

\begin{document}
	
\begin{frontmatter}
		
		
		\title{Stochastic L-system Inference from Multiple String Sequence Inputs}
		
		
		\author[1]{Jason Bernard}
		\ead{jason.bernard@usask.ca}
		\author[1]{Ian McQuillan}
		\ead{mcquillan@cs.usask.ca}

		\address[1]{Department of Computer Science, University of Saskatchewan, Saskatoon, Canada}
		
		\begin{abstract}
			Lindenmayer systems (L-systems) are a grammar system that consist of string rewriting rules. The rules replace every symbol in a string in parallel with a successor to produce the next string, and this procedure iterates. In a stochastic context-free L-system (S0L-system), every symbol may have one or more rewriting rule, each with an associated probability of selection. Properly constructed rewriting rules have been found to be useful for modeling and simulating some natural and human engineered processes where each derived string describes a step in the simulation. Typically, processes are modeled by experts who meticulously construct the rules based on measurements or domain knowledge of the process. This paper presents an automated approach to finding stochastic L-systems, given a set of string sequences as input. The implemented tool is called the Plant Model Inference Tool for S0L-systems (PMIT-S0L). PMIT-S0L is evaluated using $960$ procedurally generated S0L-systems in a test suite, which are each used to generate input strings, and PMIT-S0L is then used to infer the system from only the sequences. The evaluation shows that PMIT-S0L infers S0L-systems with up to $9$ rewriting rules each in under $12$ hours. Additionally, it is found that  $3$ sequences of strings is sufficient to find the correct original rewriting rules in $100\%$ of the cases in the test suite, and $6$ sequences of strings reduces the difference in the associated probabilities to approximately $1\%$ or less.
		\end{abstract}
		
		
		
		\begin{keyword}
			Lindenmayer systems
			\sep Plant modelling
			\sep Model inference
			\sep Stochastic simulations
			\sep Hybrid search algorithm
			
			
			
		\end{keyword}
		
	\end{frontmatter}
	
	
	\section{Introduction}\label{section:intro}
	
	In 1968, Lindenmayer \cite{lindenmayer_lsystems} proposed a grammar system later called Lindenmayer systems (L-systems) to model multicellular growth in plants. They consist of rewriting rules that are used to replace, in parallel, every symbol in a string with a word. The process of replacing all symbols with words in this manner is called a derivation step. Starting from an initial word, derivation steps iterate, thereby producing a sequence of strings. The strings can produce a visual simulation or model by interpreting the symbols as instructions, such as ``draw a line'', or ``turn left/right'' (e.g., \cite{drp_modeling_by_hand,beauty,galarreta_s0lbloodvessel,rongier_sedimentary}) with each string being one step of the temporal simulation. L-systems can be deterministic, implying that the simulation always proceeds the same way, with each string uniquely determined by the previous string; or stochastic, where the rules and derivations have a probability of occurring.
	
	L-systems of different types have been particularly successful at modelling plant growth \cite{beauty,ben_naoum_surveryLsystems}. Stochastic models have successfully modeled different plant structures and processes \cite{beauty, agu1985stochastic}, including Japanese Cypress trees \cite{nishida_k0l}. As will be discussed, the algorithm presented in this paper is domain agnostic, and so may have applications in other diverse research domains where stochastic L-systems have been successfully used. These other domains include modelling protein secondary structure \cite{danks_s0lproteinfolding}, arterial branching \cite{galarreta_s0lbloodvessel} and sedimentary formations \cite{rongier_sedimentary}. The process for finding a stochastic L-system has been described by Galarreta et al. \cite{galarreta_s0lbloodvessel} as requiring ``tedious and intricate handwork'' that could be improved by an algorithm to ``infer rules and parameters automatically from real \ldots images''. 
	
	Currently, models using S0L-systems are generated predominantly by hand by experts (e.g., \cite{beauty, galarreta_s0lbloodvessel, rongier_sedimentary, nishida_k0l}), which is time consuming. In \cite{danks_s0lproteinfolding}, automatic inference of S0L-systems was discussed specifically for the application of modeling protein folding. Their approach is domain specific as they use \textit{a priori} scientific knowledge regarding proteins as the basis for constructing the L-system's rules. Additionally, often the resulting models from existing approaches are assessed aesthetically \cite{danks_s0lproteinfolding, rongier_sedimentary, nishida_k0l}, which further reinforces that such approaches are domain specific as an aesthetic assessment can rarely (if ever) be transferred to another process. These two drawbacks hinder automatic inference.
	
	One approach for inferring an L-system from one or more temporal sequences of images is to divide the problem into two steps: 1) a segmentation of each image into the letters such that they would approximately draw the image in a simulator, 2) the inference of an L-system from the sequences of strings. This work is concerned with the second of these steps, which is called the inductive inference of an L-system. This paper presents a generalized algorithm for inductively inferring S0L-systems from one or more sequences of strings, called the Plant Model Inference Tool for Stochastic Context-free L-systems (PMIT-S0L), which despite the name is domain independent. PMIT-S0L could possibly reduce the time and effort required to produce a model of a process from imagery. Furthermore, since PMIT-S0L uses no \textit{a priori} information, it can also help reveal the mechanisms underlying a process (perhaps hidden from the images themselves), and thereby have additional scientific impact.
	
	Inductive inference was defined in the 1970's, and studied briefly in the theoretical computer science community \cite{doucet_algebra,herman}. It has recently been studied for deterministic context-free L-systems \cite{bernard_pmitml,mcquillan_poly} where it was found that all systems in a test suite of $30$ L-systems found in the literature could be inferred accurately, each in under $4$ seconds. Inference of L-systems generally was surveyed in \cite{ben_naoum_surveryLsystems}.
	
	One of the challenges with S0L-system inference, in comparison to deterministic L-systems, is the multitude of systems that may produce the sequences of strings. For PMIT-S0L, it is argued within that given a set of L-systems that can produce the string sequences, the best choice (in absence of any additional information) is an S0L-system that has the greatest probability of having produced the input strings. As such, the core concept for PMIT-S0L's algorithm is a greedy selection process that chooses rewriting rules such that after each choice, the result would be the S0L-system that locally maximizes the probability of producing the input strings. 
	
	In this paper, the implementation of PMIT-S0L is described. It uses a greedy algorithm hybridized with a search algorithm (exhaustive search and genetic algorithm are evaluated as the search algorithms). Then, an evaluation is presented for PMIT-S0L at inferring S0L-systems using $960$ procedurally generated S0L-systems. These systems were generated using statistical properties of existing L-systems in order to be equally as complex as L-systems created by experts. Procedurally generated L-systems were used since only one S0L-system could be found explicitly in the literature \cite{nishida_k0l} (other papers \cite{danks_s0lproteinfolding,galarreta_s0lbloodvessel,rongier_sedimentary} created them but do not include it in their respective paper). The limited number of stochastic L-systems in the literature is due to the difficulty in constructing them (\hspace{1sp}\cite{nishida_k0l} took an entire lengthy paper to create and justify theirs as will be described in Section \ref{section:background}), which would be dramatically improved by an automated approach. PMIT-S0L is also evaluated on this published S0L-system for Japanese Cypress \cite{nishida_k0l}. Additionally, the effects of having $M > 1$ sequences of strings as input for different values of $M$ is investigated with respect to differences between the true S0L-system and the S0L-system provided as a solution by PMIT-S0L. 
	
	Scenarios where more than one sequence are used are important as, for example, it would be possible to have imagery from many plants of the same genotype of a species (or different genotypes of a species), and to desire a stochastic model to describe the population of plants. Furthermore, it would be particularly useful to compare populations of plants by comparing the models. Additionally, stochastic L-system models that can describe a diversity of plants are useful indirectly to improve image recognition tasks. For example, Ubbens et al. \cite{drp_rosette} use a large set of synthetic L-system images	of \textit{Arabidopsis thaliana} rosettes to augment the size of a data set used to train a deep convolutional neural network for the purposes of leaf counting. This was found to improve leaf counting on real rosette images versus only training using real images.
	
	This work is an extension of the conference paper \cite{bernard_pmitsol} that had the additional restrictions of only using one sequence of strings as input, and it had a limitation with respect to not having two successors of the same symbol with one being a prefix of the other  --- the so-called \textit{prefix limitation}. This paper additionally presents methods for removing both limitations. PMIT-S0L is evaluated using a variety of performance metrics. Detailed in Section \ref{section:inferring}, the success rate describes the percentage of times that an S0L-system is found that has an equal or greater probability of producing the input than the original S0L-system has of producing the input. Thus, if the original L-system was defined as the correct L-system, then it can be possible to find L-systems that are more likely to have generated the input, and are therefore better than the L-system which actually generated the strings. The time taken to find a solution (or report that none exists) is also measured. Additional metrics are used to capture the differences between the predicted solution and the original system, both in terms of rewriting rules and probabilities. 
	
	The remainder of this paper is structured as follows. Section \ref{section:background} provides some background information on L-systems and some applications of S0L-systems. Section \ref{section:inferring} discusses some of the unique challenges of inferring stochastic L-systems, and how PMIT-S0L functions. Section \ref{section:methodology} provides the methodology for evaluating PMIT-S0L, focusing on the methods used to evaluate multiple sequences of strings as input. Section \ref{section:results} provides the results of the evaluation. Finally, Section \ref{section:conclusions} concludes the paper and discusses future directions of PMIT-S0L and inferring L-systems.
	
	\section{Background and Preliminaries}\label{section:background}

	To start, some basic formal notation is required. An alphabet is a finite set of symbols. Given an alphabet $V$, $V^*$ is the set of all words over $V$. For a word $x$, the length of $x$ is denoted by $|x|$. And for a finite set $Y$, the number of elements in $Y$ is denoted by $|Y|$.
	
	Formally, context-free L-systems are described by an ordered tuple $G = (V, X, P)$ where $V$ is an alphabet, $X$ is a finite set of strings $\{x_{1},\ldots,x_{q}\}$ where each $x_{i}$, $1 \le i \le q$ is a word in $V^*$ called an axiom (some definitions have only one axiom), and $P$ is a finite set of productions (also referred to as rewriting rules, although the term productions will be used herein). Each production is of the form $A \rightarrow \alpha$ where $A \in V$ is called the predecessor and $\alpha \in V^*$ is called the successor of the production (or a successor of $A$). The system is said to be non-erasing (also called propagating in the L-systems literature) if $\alpha$ is not the empty word for any production. The term context-free here means that the neighboring symbols in the string do not affect the selection of a successor. A derivation step is defined by, $u \Rightarrow v$, if $u=A_1\cdots A_n$, $v=\alpha_1 \cdots \alpha_n$, and $A_l \rightarrow \alpha_{l} \in P$, for $1 \le l \le n$. The system is called deterministic if $X$ only contains one string, and $P$ contains exactly one rule with each symbol in $V$; with deterministic systems, a derivation step on a string involves taking, in parallel, every symbol in the string and replacing it with its unique successor. In a stochastic L-system, every $A \in V$ has a set of one or more successors each with an associated probability of being selected, such that the sum of the associated probabilities for each $A \in V$ equals $100\%$. When performing a derivation step with a stochastic L-system, for each symbol in a string, a successor is chosen from the set of corresponding successors with the associated probability. Formally, a stochastic context-free, or S0L-system, \cite{eichorst_stochastic} is a quadruple $G=(V,X,P,p,I)$, where $(V,X,P)$ is a context-free L-system, $p$ is a function from $P$ to $(0,1]$ such that, for all $A \in V$, $\sum\limits_{A\rightarrow \alpha \in P} p(A\rightarrow \alpha) = 1$ (that describes the probability of applying a production), and $I$ is a function from $X$ to $(0,1]$ such that $\sum_{x \in X} I(x) = 1$ (that describes the probability of starting with an axiom).
		
	In \cite{eichorst_stochastic}, the authors continue with the following definitions for derivations of an S0L-system. Given $x,y$ as words in $V^*$ where $x \in X$, a derivation $d$ of $x$ to $y$ of length $m$ consists of two items:
	
	\begin{enumerate}
		\item a trace, which is a sequence of $m+1$ words $(w_{0}, \ldots, w_{m})$ such that $x = w_{0} \Rightarrow \cdots \Rightarrow w_{m} = y$,
		\item a function $\sigma$ from the set of pairs $\{(j,l) \mid 0\le j < m, 1 \le l \le |w_{j}|\}$ into $P$ such that, for $j$ from $0$ to $m-1$, if $w_{j} = A_{1} \cdots A_{|w_{j}|},A_{l} \in V$, then $w_{j+1} = \alpha_{1}\cdots\alpha_{|w_{j}|}$ where $\sigma(j,l) = (A_{l} \rightarrow \alpha_{l})$ for $l$ from $1$ to $|w_{j}|$.
	\end{enumerate}
	
	\noindent Thus, the function $\sigma$ describes the productions applied to each letter in the derivation. Given such a derivation $d$, the probability of $w_{j}$ deriving $w_{j+1}$, $P(w_{j} \Rightarrow w_{j+1}, d)$ is $\prod_{l=1}^{|w_{j}|} P(\sigma(j,l))$. Further, the probability of $d$ occurring is $P(d) = I(x) \cdot \prod_{j=0}^{m-1} P(w_{j} \Rightarrow w_{j+1},d)$. Lastly given a finite set of derivations $\{ d_1, \ldots, d_M\}$, the probability of them occurring is
	
	\begin{equation}\label{derivationprob}
	\prod_{i=1}^{M} P(d_i).
	\end{equation}
	
	As mentioned in Section \ref{section:intro}, modeling with L-systems is done by associating a meaning to each symbol (or a subset of symbols) in $V$; typically the meaning is an instruction for simulation software such as the ``virtual laboratory'' \cite{vlab}. So, a string of symbols is then taken as a sequence of instructions, and each derived word is taken as the next step in a temporal process. Symbols may have graphical and/or mechanical functions within the resulting model. A common graphical interpretation is the so-called \textit{turtle graphics} \cite{beauty}, which imagines a turtle in a 2D or 3D space with a state consisting of a position and orientation. The graphical symbols then modify the turtle's state. When the turtle moves forward, it may optionally simultaneously draw a line. For branching models, two graphical symbols (``['' and ``]'') will push and pop the turtle's state onto a stack and switch to that state. Other symbols are used to represent components or an underlying mechanism in the model. For example, a symbol may be used to represent the apex of a plant where the stem will continue to grow at the next derivation step, until it flowers; this can be modeled stochastically \cite{beauty}. 
	
\begin{figure}
		\centering
		\begin{subfigure}{.6\linewidth}
			\centering
			\includegraphics[height=4.6cm,keepaspectratio]{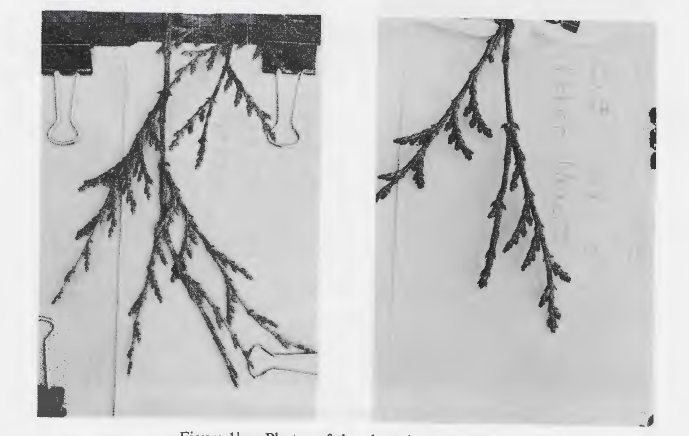}
			\caption{Observations H-4 of Japanese Cypress}
			\label{fig:1}
		\end{subfigure}
		\begin{subfigure}{.37\linewidth}
			\centering
			\includegraphics[height=4.6cm,keepaspectratio]{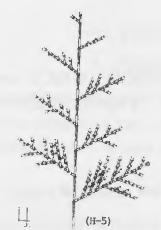}
			\caption{Image produced by an S0L-system similiar to observations H-4}
			\label{fig:2}
		\end{subfigure}
		\\
		\begin{subfigure}{.6\linewidth}
			\centering
			\includegraphics[height=4.6cm,keepaspectratio]{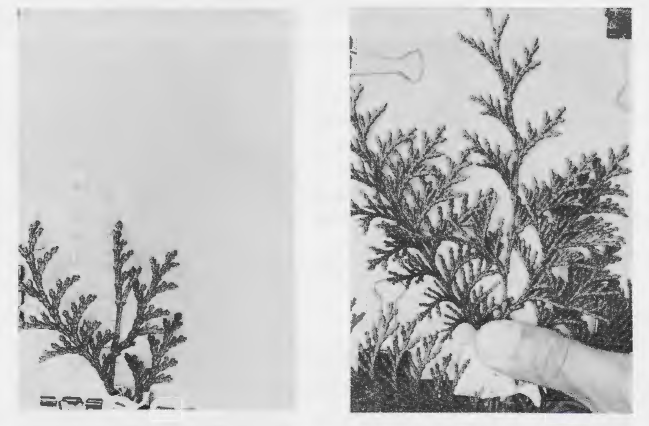}
			\caption{Observations H-6 of Japanese Cypress}
			\label{fig:3}
		\end{subfigure}
		\begin{subfigure}{.37\linewidth}
			\centering
			\includegraphics[height=4.6cm,keepaspectratio]{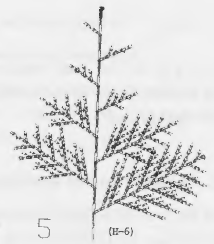}
			\caption{Image produced by an S0L-system similiar to observations H-6}
			\label{fig:4}
		\end{subfigure}
		\caption{Images from \cite{nishida_k0l}, reprinted with permission of Kyoto University.}
	\end{figure}

	
	Nishida \cite{nishida_k0l} investigated using S0L-systems to model Japanese Cypress trees. He did not use any \textit{a priori} biological knowledge of Japanese Cypress, and instead used a process of converting imagery to segments, and then to an L-system. Importantly, this is the same process as our main goal (images to segments to L-systems, with PMIT-S0L being useful for the second step), except their work was done completely manually. In the paper, the meticulous process is described of segmenting images of Japanese Cypress by hand. The segments are then mapped to letters of an alphabet, with successors and associated selection probabilities picked for the segments such that they would reproduce the appropriate segmentation in the next image. The result is an S0L-system with $23$ symbols and a total of $42$ productions (shown in Table SD1 in the supplementary materials). The system produces imagery similar to the photos of the Japanese Cypress, as seen in Figures \ref{fig:1} to \ref{fig:4} \cite{nishida_k0l}. This shows that the goal of automatic inference in exactly this fashion is an exciting opportunity, as doing so manually requires a huge amount of effort.
	
	\section{Inferring S0L-Systems using Greedy Algorithm}\label{section:inferring}
	
	This section will describe the methodology of PMIT-S0L. For the remainder of this paper let $\rho = \{\rho_{1}, \ldots ,\rho_{M}\}$ be the input, where each $\rho_{i}$ is a finite sequence of strings over $V$. Furthermore, let $\rho_{i} = (w_{i,1}, \ldots ,w_{i,m_{i}})$, $1 \le i \le M$ (each $w_{i,j}$ is a string in $V^{*}$). In this paper, we assume $m_{1} = \cdots = m_{M}$ (all sequences can be truncated to have the same number of strings), and we denote this size by $m$.
	Inferring an L-system can be stated as outputting some L-system $G$ (or reporting that none exists) that could produce all the sequences in $\rho$; that is, $G$ has a derivation with a trace of $\rho_{i}$, for each $i$, $1 \le i \le M$. In this case, $G$ is said to be compatible with $\rho$. To do this, the algorithm scans each sequence $\rho_{i}$ in $\rho$ and attempts to determine a derivation, which has a trace of $\rho_{i}$ in a word-by-word fashion. Each time it determines part of a derivation (e.g., that some specific occurrence of a letter in $w_{i,j}$ produces the subword between two positions of $w_{i,j+1}$), it adds this production to the current list of productions if it has not already been added.
	
	\subsection{Complications with Stochastic Inference}
	
	If a compatible L-system does exist, then it must be found in the space of all L-systems, and hence it is at least possible to search for one. PMIT-D0L \cite{bernard_pmitdv3,bernard_pmitml} is an existing tool for inferring deterministic context-free L-systems (D0L-systems). The first PMIT-D0L implementation \cite{bernard_pmitdv3} used genetic algorithm, and searched for successors as an ordered sequence of symbols in a search space that was pruned using mathematical properties based on necessary conditions of L-systems. A significant improvement was made when it was recognized that every successor of a symbol $A \in V$ must be a subword of every word directly after one where $A$ occurs (for a deterministic L-system $M = 1$ is enough since the word produced at each step is uniquely determined), and that searching for an ordered sequence of symbols could be replaced by searching for successor lengths for each letter of the alphabet \cite{bernard_pmitml,mcquillan_poly}. With each possible solution consisting of a successor length for each $A \in V$, it is possible to scan every symbol in each word in $\rho$ starting from the first word, and take the subword of the associated length as the successor. When a scan can be completed without any inconsistencies, then the resulting successors are compatible with $\rho$. This is referred to as the \textit{scanning process}. This approach works in the deterministic case because any information deduced about the successor for any instance of $A$ in $\rho$ must be true for every $A$ that appears in every word in $\rho$, and therefore this search space has only $|V|$ dimensions. 
	
	For stochastic L-systems, while it still true that every $A \in V$ must produce a subword of the next string and hence some kind of encoding scheme with successor lengths is possible, it is no longer true that deducing a fact about some instance of $A$ is true for all other instances of $A$. Indeed, with S0L-systems, different instances of each $A$ (of each word of each sequence) in $\rho$ can produce different successors. For $\rho_{i} = (w_{i,1}, \ldots, w_{i,m})$, then since every instance of a symbol can produce a different successor, the most intuitive solution space defines at least one dimension (for example, using the successor length encoding scheme from \cite{bernard_pmitml,mcquillan_poly}) for every symbol in each word of each sequence to produce a search space with $N$ dimensions, where $N = \sum_{i=1}^M \sum_{j=1}^{m} |w_{i,j}|$. This is clearly intractable as an extra dimension is needed for every additional symbol in $\rho$. Furthermore, reducing the bounds for the dimensions (e.g., upper and lower bounds on successor length) is very difficult since there is little information upon which to deduce such bounds (due to the aforementioned issue that every instance of a symbol can produce a new successor). The lower bounds of each successor length is $1$ (productions are assumed to be non-erasing in this paper) and the upper bound for an instance of $A$ would be the length of the next word minus the number of symbols in the previous word (since every symbol produces at least one symbol) directly after one where that $A$ occurs. In summary, to search for an S0L-system with the entire search space in a similar fashion to what has been done for D0L-systems, requires both a number and scale to the dimensions that would be too large to allow for a practical search time. 
	
	A further complication when inferring S0L-systems is there exist a multitude of possible S0L-systems that are compatible with $\rho$. By taking each pair of consecutive words $w_{i,j}$,$w_{i,j+1}$ in every $\rho_{i}$ of $\rho$, then every way of dividing $w_{i,j+1}$ into $|w_{i,j}|$ subwords can be used to create productions that lead to a compatible S0L-system. However, one crucial and distinguishing property for any of these candidate solutions is the probability that it produces $\rho$. A best possible solution compatible with $\rho$ is desired rather than an arbitrary solution.

	\subsection{Methodology with PMIT-S0L}
	
	PMIT-S0L infers an S0L-system based on $\rho$ by selecting successors, such that each choice locally maximizes the probability of producing the words of $\rho$ scanned so far. This paper investigates inferring an S0L-system by scanning the words symbol-by-symbol, and choosing each successor based on a successor length (similarly to PMIT-D0L in \cite{bernard_pmitml,mcquillan_poly}) by preferring successors that have been previously selected using a \textit{greedy algorithm}. 
	
	To determine an S0L-system, an alphabet, axioms, and rewriting rules with associated probabilities need to be predicted. The alphabet $V$ is found easily by recording every unique symbol in $\rho$, and so we henceforth assume it is known. For the S0L-system, multiple axioms are assumed. In particular, the set of all axioms $X$ is assumed to be the set of all of the first words of sequences in $\rho$. We do not attempt to infer the probability of starting with each axiom (which could just be calculated as the number of times each axiom is the first word of a sequence in $\rho$ divided by $M$). Also, computing a reduced number or a single axiom is an area for future investigation. The process for inferring the rewriting rules and their probabilities is more involved and is described in remainder of this section.
	
	As mentioned earlier, for any $\rho$, there are a multitude of possible compatible S0L-systems, but a best solution is a S0L-system with the greatest probability of producing $\rho$ \cite{bernard_pmitsol}. In absence of any additional information (e.g., \textit{a priori} knowledge that might lead to a different choice), the S0L-system with the greatest computed probability is said to have maximum parsimony. For example, Table \ref{table:method1} shows two abstracted S0L-systems, where in parentheses is the number of times that each successor was applied to produce a sequence of strings in the two different derivations. The ``Odds'' column shows the computed probability of the derivation occurring. It can be seen that the first S0L-system has a greater probability of producing the sequence of strings, and so should be preferred as the solution. The probability that an S0L-systems produced $\rho$ is increased when one or a few successors have a high probability, as opposed to an equal distribution across all of the successors. This mathematical property provides the guidance for a greedy algorithm to infer an S0L-system using the process described next.
	
	\begin{table}
		\centering
		\begin{tabular}{|c|c|}
			\hline
			Successors & Odds \\ \hline
			$A \rightarrow \alpha_{1}: 90\% (9)$ & \multirow{2}{*}{$0.9^{9} \times 0.1^{1} = 3.87\%$} \\ 
			$A \rightarrow \alpha_{2}: 10\% (1)$ & \\ \hline
			$A \rightarrow \alpha_{3}: 50\% (5)$ & \multirow{2}{*}{$0.5^{5} \times 0.5^{5} = 0.097\%$} \\ 
			$A \rightarrow \alpha_{4}: 50\% (5)$ & \\ \hline
		\end{tabular}
		\caption{Two abstracted S0L-systems with odds that it would generate a specific set of strings.}
		\label{table:method1}
	\end{table}
	
	Conceptually, the core greedy algorithm component is relatively straightforward. Suppose that a list of successors for each $A \in V$ is being maintained including a count of how often each successor has been selected. For each word $w_{i,j}$ with $j$ from $1$ to $m$, the algorithm will partition $w_{i,j+1}$ into $|w_{i,j}|$ subwords by scanning each letter $A$ in order from left-to-right as follows: If $A$ is the rightmost symbol of $w_{i,j}$, then pick the successor to produce everything that remains in $w_{i,j+1}$ (the parts of $w_{i,j+1}$ that have not been matched in the derivation). If $A$ is not the rightmost symbol, then pick the successor, if one exists, from the current list for $A$ out of those that match the next symbols of $w_{i,j+1}$ that has been selected most often so far. One issue with this algorithm is that early on, especially for the first few symbols scanned, the list of successors for each $A \in V$ is (or is near) empty or none match, and so the greedy process is not able to make a choice from the existing list. Sometimes in the literature, this type of problem is resolved using a random forest algorithm \cite{breiman2001random}, but this was found to not work well in this instance (discussed below). A search algorithm was used in these instances to pick successors when nothing in the list is matched.
	
	For this, the algorithm keeps a vector $y$ of $N$ non-negative integers. When there is no successor in the current list for $A$ that matches the next symbols in $w_{i,j+1}$, the algorithm retrieves the next unused integer from $y$, $k$ say, and then a successor is selected using the next $k$ symbols from $w_{i,j+1}$. To find $y$ (a single sequence for all letters), a search algorithm is used (this search process will be described in Subsection \ref{section:data}). However, this process raises a new question: how many successor length choices are needed; i.e., what should be $N$?
	
	The best value for $N$ is the number of distinct successors in the best S0L-system, which is difficult to accurately predict in advance. However, if a guess at $N$ is made, then either the guessed $N$ will be exactly right, too low, or too high. If $N$ is exactly right, then there are no issues. If $N$ is too high, then so too will be the execution time, i.e., the search space becomes larger than necessary. If $N$ is too low, then it is possible that the algorithm may reach a point where the greedy algorithm cannot make an appropriate choice and there is no element left in $y$. In this case, then either the search can be restarted with a higher value of $N$ or $y$ could be extended by one dimension. However, some solutions will encounter this issue repeatedly, thus requiring many extensions leading to an increased execution time. In practice, simply extending each time as needed is probably intractable; e.g., there would be candidate solutions that would required hundreds or thousands of additional dimensions. A balance can be achieved by having the implementation extend $y$ by a limited additional number of times per word of $\rho$ (PMIT-S0L uses a limit of one), which can help find correct solutions while keeping the expansion of the search space reasonable. If more than the chosen limit is needed, then a higher value of $N$ is required.
	
	Lastly, the process will terminate with error when none of the methods above can produce a successor for the current symbol $A$ (this only occurred when $N$ was too small in our tests). In this case, $N$ needs to be incremented; however, it is unclear what is an optimal value for the increment. A small increment for $N$ minimizes the growth of the search space, but increases the chance that the search will fail again. A large increment for $N$ increases the chance that the subsequent search will succeed, but will possibly make the search space larger than necessary (certainly an increment of $1$ will eventually find the correct $N$). In our experiments, a value of $N$ that is too low is simply considered a failed state in order to better understand its effect on the time taken to complete. The value of $N$ can be increased manually or automatically if desired. 
	
	\begin{envirolist}\label{procedure:1}
		In summary, for a given integer sequence $y$, the successor selection process works by choosing the first applicable rule when scanning each character in $\rho$. Let $z := 1$ be a programming variable.
		\begin{enumerate}
			\item If scanning the last symbol of the current word, then select all remaining symbols in the produced word,
			\item An existing successor in the list for $A$ matches the next symbols in $w_{i,j+1}$; the most frequently chosen thus far is selected greedily,
			\item Build a successor of length $y_{z}$, and increase $z$ by one,
			\item Terminate with error.
		\end{enumerate}
	\end{envirolist}
	
	Given an appropriately chosen $y$, this process can determine a compatible S0L-system. Furthermore, notice, that this procedure is determining the derivations associated with traces in $\rho$. However, the solution need not be optimal (the greedy choice might not lead to the best solution).
	
	\subsection{Searching for Successor Lengths}\label{section:searching}
	
	Next, the method for determining the integer sequences $y$ is described. Two algorithms, standard genetic algorithm (SGA) and exhaustive search (ES), are evaluated. For both, a literal encoding \cite{back_geneticalgorithm} is used, so a search space is constructed of $N$ integer dimensions with bounds from $1$ to $10$ (as discussed and justified later in Section \ref{section:data}, the maximum number of symbols in each successor is assumed to be $10$ although this could be increased). For each sequence $y$ searched, Procedure \ref{procedure:1} is used to predict an S0L-system $G$ compatible with $\rho$, and the derivations corresponding to each trace in $\rho$. The fitness value of $y$ is defined to be the probability of these derivations occurring with the traces in $\rho$, which is given in Equation (\ref{derivationprob}). Algorithm \ref{algo:1} shows the pseudocode for PMIT-S0L.
	
	\begin{algorithm}
		\SetAlgoLined
		\KwData{A set of $M$ sequences of strings over $V$ called $\rho$, a search algorithm $S$}
		\KwResult{$B$, the S0L-system with the greatest probability of producing $\rho$ of those searched; or reports that no compatible S0L-system was found}
		$T := false$ // variable to track if the algorithm should terminate\;
		$B$ // stores the best solution found thus far\;
		$F(B) := 0$ // stores the fitness of $B$\;
		
		\Repeat{$T = true$} 
		{
			$S.iterate()$ // performs one iteration appropriate for the search algorithm\;
			\ForEach{$y \in S.population$}
			{
				// Produce an S0L-systems $G$ using the rules in Procedure \ref{procedure:1}\;
				$(G,fitness) := Produce(y)$\;
				
				\uIf{$fitness > F(B)$}{
					$B := G$\; 
					$F(B) := fitness$\;				
				}
			}
			$T := S.terminate()$ // determines if $S$ should terminate in an algorithm specific way\;
		}
		
		\caption{This pseudocode describes the process for searching for an S0L-system with a higher probability of generating $\rho$.}\label{algo:1}
	\end{algorithm}
	
	When the search technique is set to SGA, it uses roulette wheel selection, uniform crossover, uniform mutation, and elite survival operators \cite{back_geneticalgorithm}. An SGA with simple operators is selected as little is known about the search space and an SGA provides easily tunable mechanisms for balancing exploration and exploitation. Briefly, an SGA works by the following steps \cite{back_geneticalgorithm}. It consists of a population of $P$ genomes, where each genome in this case is a candidate vector $y$ consisting of $N$ genes. The initial population is produced randomly. The main processing loop of an SGA performs four steps: selection, crossover, mutation, and survival. In the selection step, $P/2$ pairs of genomes are selected from the population. The crossover step takes each pair and randomly swaps zero or more genes between them, and since there are $P/2$ pairs, this results in $P$ new genomes. Each new genome is then mutated by randomly changing zero or more genes to a new value. The survival step merges the initial and new populations. Each member of the population is assessed a fitness value as described above (using Procedure \ref*{procedure:1} with $y$ from the population), and the population is sorted by fitness value. The top $P$ genomes are preserved and the remainder are culled. With respect to Algorithm \ref{algo:1}, $S.iterate()$ for an SGA corresponds to performing the selection, crossover, mutation, and survival steps. $S.population$ is the entire population of genomes. The SGA terminates by convergence detection ($S.terminate()$ in Algorithm \ref{algo:1}), which functions as follows. Termination occurs as follows: every time a new best solution is found, a variable $Gen$ records the number of the current generation. If $Gen$ additional iterations occur without a new best solution being found, then the SGA terminates. 
	
	To optimize the control parameters of the SGA, a hyperparameter search was done \cite{hyperparameter} using a random search with $16$ trials. As a result of the hyperparameter search, the control parameters were set as follows: population size of $50$, crossover weight of $0.9$, and a mutation weight of $0.01$. 
	
	In contrast, the ES simply iterates through all possibilities until it terminates, keeping track of the S0L-system with the highest fitness value. Since later dimensions are often unused, it would be preferred for these dimensions to be searched last, i.e., to search deeply into the leading dimensions, hence a depth first search is used. $S.iterate$ corresponds to searching one step deeper. $S.population$ is the number of candidates solutions in this iteration, which is always $1$ with ES. $S.terminate$ returns $true$ when there are no more nodes to search.
	
	\subsection{The Prefix Limitation}\label{section:1}
	
	In the conference paper \cite{bernard_pmitsol}, this earlier version of PMIT-S0L had an identified limitation, when for at least one $A \in V$, there are two or more successors, and one successor is a prefix of the other \cite{bernard_pmitsol}. Let $u,v$ be two successors of $A$, such that $u = vx$, and $u,v,x$ are words over $V$. In this case, if the shorter successor ($u$) is encountered first in $\rho$, then assuming all other successors are correct, for all future instances of $A$, the greedy choice (rule $\#2$ of Procedure \ref{procedure:1}) will always select $u$ as the successor as the next $|u|$ symbols can be positively associated to $A$. This effect is shown in Example \ref{example:1} below. Since the introduction of PMIT-S0L \cite{bernard_pmitsol}, a technique has been found to often correct this limitation, described next, although it is only used with ES and not with the SGA (due to complications to be discussed).
	
	\begin{example}\label{example:1}
		Let $w_{1} = AAA$	and $w_{2} = AAAAAABBB$. Suppose that the successors for $A$ in the original system are $AAA \thinspace; AAAB \thinspace ; BB$ (with some associated probabilities that are not important for the example). Finally, assume $N = 2$ (the length of $y$), and the search algorithm has a candidate solution with the value ${3,4}$. The first $A$ will be assigned the successor $AAA$ based on the value found by the search algorithm (rule $\#3$). The second $A$ will be also assigned $AAA$ based on the greedy choice (rule $\#2$), but this decision is incorrect as the desired choice is $AAAB$. Finally, the third $A$ will also have the wrong successor of $BBB$ (chosen by rule $\#1$).
	\end{example}
	
	In this example, in order to find the correct successor, the greedy choice (rule \#2) needs to be ignored and instead to use the $2^{nd}$ value in the search space ($4$) should be used to find the successor $AAAB$. Instead, $y$ is modified to be of the form $y=(t_{1},y_{1},t_{2},y_{2},\ldots , t_{N},y_{N})$. Procedure \ref{procedure:1} is modified so that after using $y_{z}$ for the length in rule $3$, it only allows at most $t_{z+1}$ greedy choices before forcing it to not apply any more greedy choices (skip rule 2 and apply rule 3). Consider the following example:
		
	
	
	\begin{example}\label{example:1b}
		Suppose the first few values of $y=(t_{1},y_{1},t_{2},y_{2},t_{3},y_{3},\ldots)$ are $(0,3,3,4,1,5,\ldots)$. Then, the first five successors have been found by:
		
		\begin{enumerate}
			\item Build a successor of length $3$ since $y_{1} = 3$
			\item A greedy choice
			\item A greedy choice
			\item A greedy choice
			\item Stop allowing greedy choices since $t_{2} = 3$
			\item Build a successor of length $4$ since $y_{2}=4$)
			\item A greedy choice
			\item Stop allowing greedy choices since $t_{3} = 1$
			\item Build a successor of length $5$ since $y_{3} = 5$
		\end{enumerate}
	\end{example}

	In this way, the search procedure used to produce $y$ also dictates exactly when new productions should be created according to $y$, even if a greedy choice can be applied. Furthermore, the modified search space can be pruned if the end of $\rho$ is reached and not all elements in $y$ have been used up to $y_{N}$ (other vectors $y$ with different values in the unused parts do not need to be considered). Similarly, if the $t$ values leads to an incompatibility, then certain vectors can be pruned. Since it is easier to prune these values with ES, adjusting PMIT-S0L to remove the prefix limitation requires the use of ES.

	Algorithmically, PMIT-S0L has a Boolean control parameter called \textit{prefix limitation} (henceforth, $PL$) that controls whether it uses this alternate procedure to address the limitation. Where relevant for discussing differences (e.g., Results) PMIT-S0L+PL indicates that the prefix limitation variable it set to \texttt{true}, while PMIT-S0L-PL indicates that it is set to \texttt{false}.
	
	It is acknowledged that while this technique can address the prefix limitation, it is somewhat inefficient (shown in Section \ref{section:results}). Finding a more efficient technique to address this limitation is an area for future investigation.
	
	\section{Evaluation Methodology}\label{section:methodology}
	
	This section describes the experimentation used to evaluate PMIT-S0L at successfully inferring (parsimonious) compatible S0L-systems for one or more sequences of strings. This section starts by discussing the metrics used to measure the success of PMIT-S0L. This is followed by a description of the procedural generation process used to produce the test set.
	
	\subsection{Performance Metrics}
	
	The metrics used to evaluate PMIT-S0L are described next. In these metrics, the \textit{original system} is the hidden L-system that generated the input strings, and the \textit{candidate} is the predicted L-system. While the original L-system can be thought of as the ground truth, it is actually possible to find an L-system that is more likely to generate the input $\rho$ than the original L-system. Therefore, the main goal is to find one with the highest probability of generating $\rho$ rather than the original. This can occur when the original L-system produces $\rho$ by chance that is unlikely for it. As the number of string sequences in $\rho$ increases, the less likely this should be. Hence, the main research question of this work is to investigate the effect of inferring a candidate L-system when multiple sequences of strings are used as input.
	
	Several measures are used to assess accuracy, which is necessary in order to properly capture different ways in which S0L-systems can differ: success rate (SR), mean time to solve (MTTS), weighted true positive - system to candidate (WTP-H2C), weighted true positive - candidate to system (WTP-C2S), probability error ($e$), maximum successor difference (\textit{diffmax}), and successor difference rate (\textit{diffrate}), which will be described next.
	
	Success for PMIT-S0L is defined as inferring an S0L-system that is compatible with $\rho$ that has equal to or greater probability of producing the sequence(s) than the original system (finding an L-system that is slightly worse than the original would therefore be classified as not successful with this stringent measure). Thus, success rate is the percentage of experiments for a set of control parameters that successfully finds such an S0L-system.
	
	MTTS is the amount of time required by PMIT-S0L to find a candidate system, whether successful or not. Execution time is limited to $12$ hours to keep overall experimental times practical (in practice, a user may be willing to wait longer for a result). All timings were captured using a single core of an Intel 4770 @ 3.4 GHz with 12 GB of RAM on Windows 10.
	
	Within the context of this work, a true positive is defined as a successor that is in both the original L-system and the candidate L-system regardless of any difference in their associated probability. One could also certainly define the following: a false positive as a successor that is in the candidate L-system but not in the original, and a false negative is the opposite. It would be possible to simply count the true positives, false positives and false negatives with these definitions. However, there are two issues: first, the terms false positive and false negative implies that original L-system is the better L-system, which might not be the case. Second, only counting successors ignores the probabilities and successors with a higher associated probability are more important toward reproducing a process \textit{in silico}. Hence, the measures listed next are used instead, and they are weighted to favor those with a higher associated probability. 
	
	\begin{enumerate}
		\item Weighted True Positive - System to Candidate (WTP-S2C) is the sum of associated probabilities for true positive successors in the original system divided by $|V|$.
		\item Weighted True Positive - Candidate to System (WTP-C2S) is the sum of associated probabilities for true positive successors in the candidate system divided by $|V|$.
	\end{enumerate}
	
	\noindent Ideally, the weighted true positive values above should be 1.00 indicating a perfect match. When a match between the hidden and candidate L-systems is imperfect, greater values generally indicate that successors with higher associated probabilities have been matched.
	
	Probability error ($e$) is calculated by taking each production in the original system; if that production is also in the predicted system, then add the absolute difference of the two probabilities; if the production is not present, then add the probability of it occurring (in the original system). Once all productions have been examined, the total is divided by the number of symbols in the alphabet. Thus, this metric is measuring how different the predicted L-system is from the original even taking into account the differences in probabilities of the rewriting rules. It is an important metric for the accuracy of inferring S0L-systems.
	
	While it is argued that successors with a higher associated probability are more important towards successfully simulating a process, the measures maximum successor difference (\textit{diffmax}), and successor difference rate (\textit{diffrate}) provide an alternative viewpoint to the weighted metrics above on how well the candidate and hidden system match. The maximum successor difference is the largest count of the number of successors that are in the hidden L-system but are not in the candidate L-system across the experiments. While this could be averaged over the number of productions, the intent of this measure is to examine if the number of extra or missing successors varies with $M$ (as opposed to by number of productions and $M$). The successor difference rate is the percentage of experiments where the successor difference was not zero. 
	
	\subsection{Data}\label{section:data}
	
	Since there are very few published S0L-systems in the literature, the test cases for PMIT-S0L are mainly procedurally generated described as follows.
	
	The procedural generation is based on observations of L-systems found in the University of Calgary’s virtual laboratory \cite{algorithmicbotany}. The observations focused on alphabet size, successor rule length, the number of successors per symbol, and the number of distinct letters in successors. Based on the observations, the alphabet size is between $2$ and $9$ symbols, the successor length may vary from $1$ to $10$, each symbol may have between $1$ and $3$ successors, and each successor has $1$ to $5$ distinct symbols. Graphical symbols (that are part of the turtle interpretation) are not used explicitly. Therefore, our experimental results are reflective of the symbols being totally arbitrary without any assumptions on the alphabet.
		
	Alphabet size is not explicitly picked, but instead is based on a value $S$ that is the total number of successors. Symbols are then iteratively added from ($A,B,C,\ldots,K$) with a randomly selected number of successors for each symbol added until $S$ is reached. Each symbol has a $50\%$ chance of having $1$ successor, a $40\%$ chance to have $2$ successors, and a $10\%$ chance to have $3$ successors. If a number of successors for a symbol is selected such that $S$ would be exceeded, then it is reduced to ensure that this does not happen. It is possible by chance that all symbols would have $1$ successor, which would be a deterministic L-system. If this occurs, then two symbols from $V$ are picked randomly; e.g. $A$ and $B$. The symbol $A$ is given another successor, and $B$ is removed ensuring the total number of successors remains correct, and that all produced systems are stochastic, but not deterministic. 
	
	When a symbol has $2$ or more successors, the associated probability ($p$) for each successor is found by iterating over the number of successors, and selecting a random value between an upper $u$ and lower bound $l$, which are programming variables. The upper and lower bounds are initialized such that $u := 100\% - (n-1)$ and $l := 1\%$. After each iteration, $u := u - p$. The last successor has $p := u$. The next step is to construct a word for each successor over $V$.
	
	To determine the length of the successor, the following probability distribution is used to determine each successor’s length (expressed as $p$ - $l$, where $p$ is the chance the successor has length $l$): $4\%$ - 1, $16\%$ - 2, $20\%$ - 3, $20\%$  - 4,$20\%$ - 5, $16\%$ - 6, $4\%$ - 7, $2\%$ - 1, $1\%$ - 1, and $1\%$ - 10. Finally, the probability distribution for the number of distinct symbols is evenly divided from $1$ to $5$ ($20\%$ chance each). Random symbols are then selected from $V$ until the successor length is reached.
	
	Using the procedure above, three data sets are generated to evaluate PMIT-S0L.
	
	The first and second data set are called $DS_{PL}$ (data set prefix limitation) and $DS_{nPL}$ (data set no prefix limitation) respectively and both enforce that $M=1$ to isolate any effects from higher values of $M$. $DS_{PL}$ furthermore enforces that all produced systems have no cases where $A \rightarrow uv$ and $A \rightarrow u$ for any $A \in V$, where $u$ and $v$ are words over $V$; i.e., no common prefix for two successors of a symbol $A$. $DS_{nPL}$ has no restriction on prefixes. The intent of these data sets is to allow an evaluation of the effect of addressing the prefix limitation, so when PMIT-S0L is executed on $DS_{PL}$ the vector $y$ produced contains only successor length values. For $DS_{PL}$, $S$ was iterated from $3$ to $10$, $60$ L-systems were generated for each value of $S$, and each experiment was conducted twice. For $DS_{nPL}$, $S$ was iterated from $3$ to $9$ (only because it was clear that PMIT-S0L would timeout with $S=10$), $60$ L-systems were generated for each value of $S$, and each experiment was conducted twice.
	
	To test the effect of using different numbers of sequences of strings $M$, an S0L-system $G$ is generated using the process described above with the following parameters. $S$ (total number of successors) is selected as a random value from $3$ to $9$. The lower bound of $3$ is the lowest possible value for and S0L-system with $|V| \ge 2$, and $|V| = 1$ is not considered as it is uninteresting. The upper bound was determined by the experiments with $SD_{nPL}$ described above. For each generated S0L-system, it generates input sequences for each $M$ iterated from $1$ to $10$. For any single experiment, when $M > 1$, the exact same sequence of strings is not permitted to be generated; if this occurs, a new sequence of strings is generated until it differs from all of the sequences produced so far. Sixty S0L-systems were generated in this fashion, and for each combination of $G$ and $M$, the experiment was done twice (for a total of $1,200$ experiments). The dataset for these experiments is denoted as $DS_{vM}$ (data set varying $M$).

	\section{Results and Discussion}\label{section:results}
	
	This section provides the results of the evaluation of PMIT-S0L. First, the evaluation of PMIT-S0L on the procedurally generated S0L-systems (the main result) is discussed, including the effects of addressing the so-called \textit{prefix limitation} (previously described in Section \ref{section:1}). Afterwards, observations regarding the inference of the L-system for Japanese Cypress trees are described. The variations of PMIT-S0L executing with and without the prefix limitation process are denoted as PMIT-S0L+PL and PMIT-S0L-PL respectively when needed for clarity.
	
	A preliminary investigation was done using only greedy algorithm, then random forest, before building a hybrid greedy algorithm with a simple genetic algorithm (SGA), and exhaustive search (ES). As they were preliminary, the results of these experiments can be found in Table SD2 of the supplementary information; however, in summary both the greedy-only algorithm and the random forest were found to be ineffective relative to the hybrid algorithm. Therefore, the remainder of this section focuses on the hybrid algorithms.
	
\begin{figure}
	\centering
	\begin{subfigure}{.48\textwidth}
	\centering
	\includegraphics[width=0.98 \linewidth]{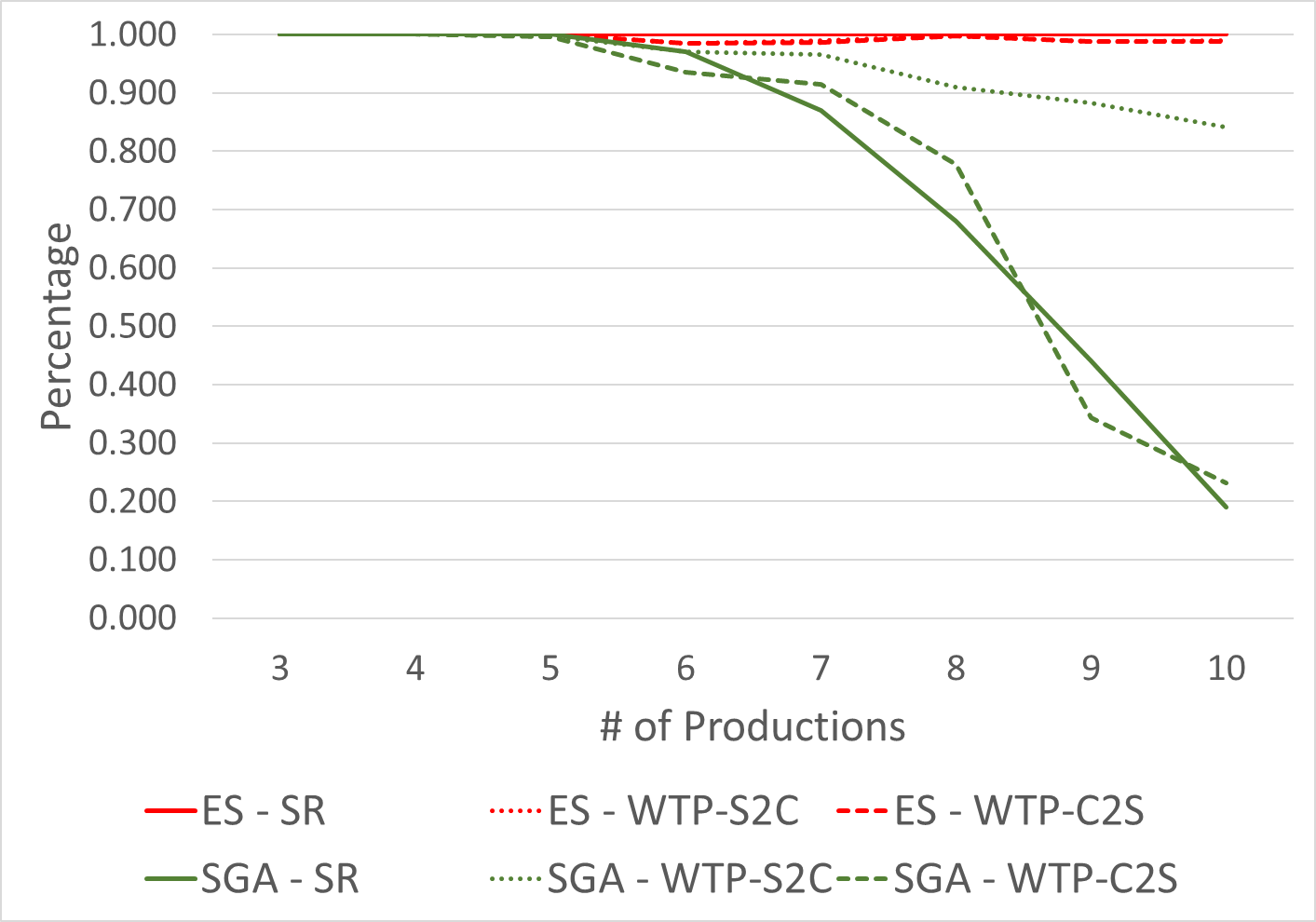}
	\label{fig:pmits0lresult1a}
	\end{subfigure}
	\begin{subfigure}{.48\textwidth}
	\centering
	\includegraphics[width=0.98 \linewidth]{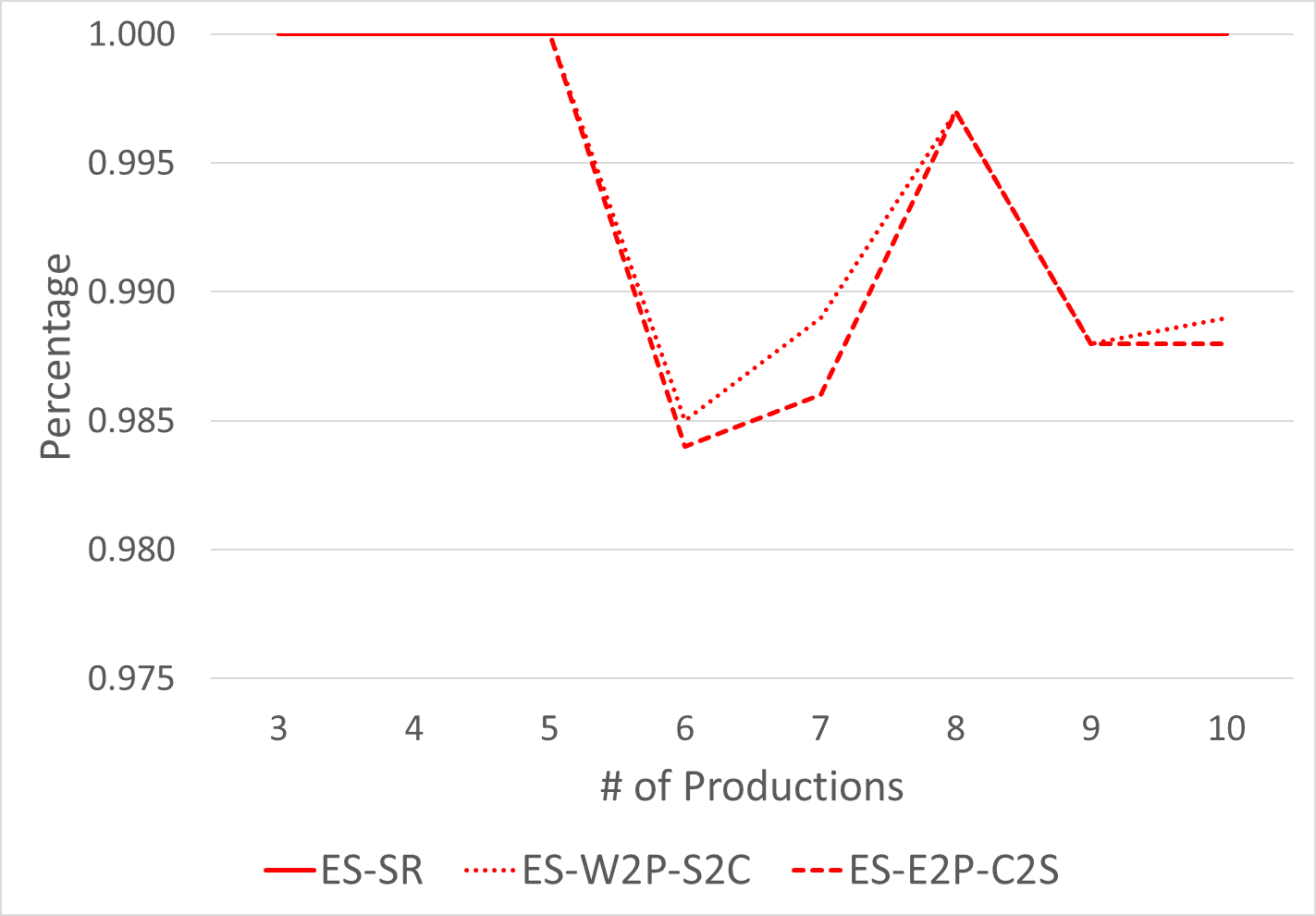}
	\label{fig:pmits0lresult1b}	
	\end{subfigure}
		\caption{On the left, the measures of accuracy SR, WT2-S2C, WTP-2CS for PMIT-S0L-PL using ES (red) and SGA (green) with data set $DS_{PL}$ are plotted against the number of productions. As it is difficult to see the three lines for ES, a zoomed view is shown on the right.}
	\label{fig:pmits0lresult1}
\end{figure}

	It was found that PMIT-S0L-PL is able to reliably infer S0L-systems with up to $10$ successors when $M=1$ (shown in Figure \ref{fig:pmits0lresult1}). However, some minor variations were noticed between the candidate solutions and the original S0L-systems shown on the two WTP lines. While SGA was not as successful as exhaustive search, it is much faster, peaking at about $2$ minutes for $10$ successors (see Table SD3, time for ES described below). A fitness landscape analysis was completed; however, there were no characteristics to the search space that suggested a particular way forward. Thus, an avenue for future investigation is to investigate the search space more deeply to find ways to avoid the use of ES for reasons of performance. 
	
\begin{figure}
	\centering
\includegraphics[width=0.90 \linewidth]{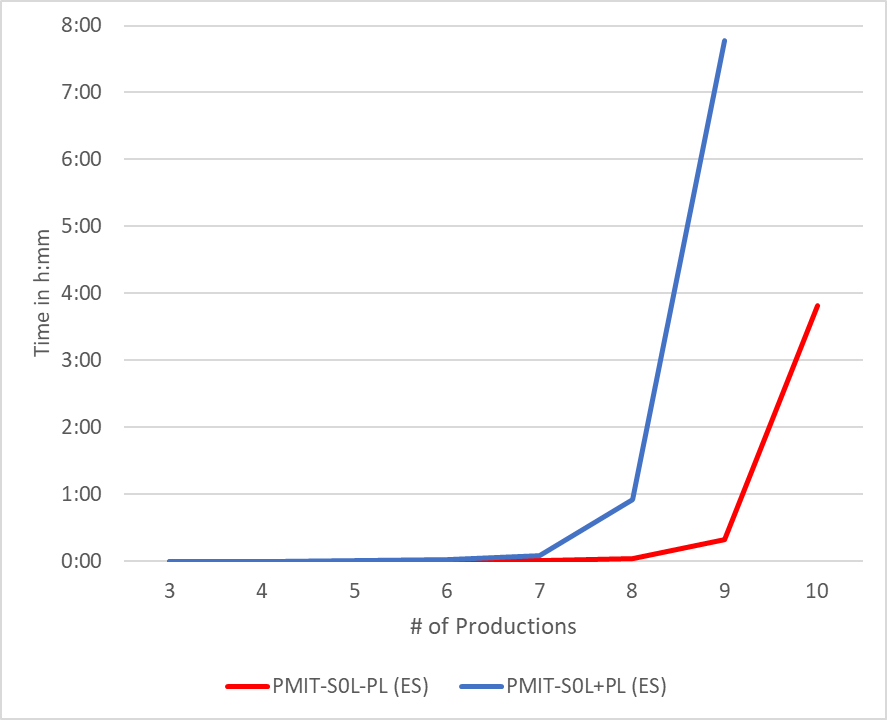}
	\caption{A comparison of MTTS versus the number of productions for PMIT-S0L-PL (red) and PMIT-S0L+PL (blue).}
	\label{fig:pmits0lresult3}
\end{figure}

	The effect on MTTS on inferring S0L-systems with and without the prefix limitation process enabled is shown Figure \ref{fig:pmits0lresult3} (the other metrics were not significantly effected so these are included in the Table SD4). It shows that PMIT-S0L+PL is slower than the other variants. It is difficult to objectively say whether the time increase is worth the ability to infer S0L-systems where a symbol $A$ has a successor that is a prefix of another, as it is unclear how often this occurs in practice. Subjectively, considering the possibility that the encoding scheme described herein to address the prefix limitation had the potential to be extremely large, the time increase seems reasonable. This suggests that, at least for the procedurally generated data sets, the method for removing the prefix limitation with pruning is effective. 
	
\begin{figure}
	\centering
	\includegraphics[width=0.95\linewidth]{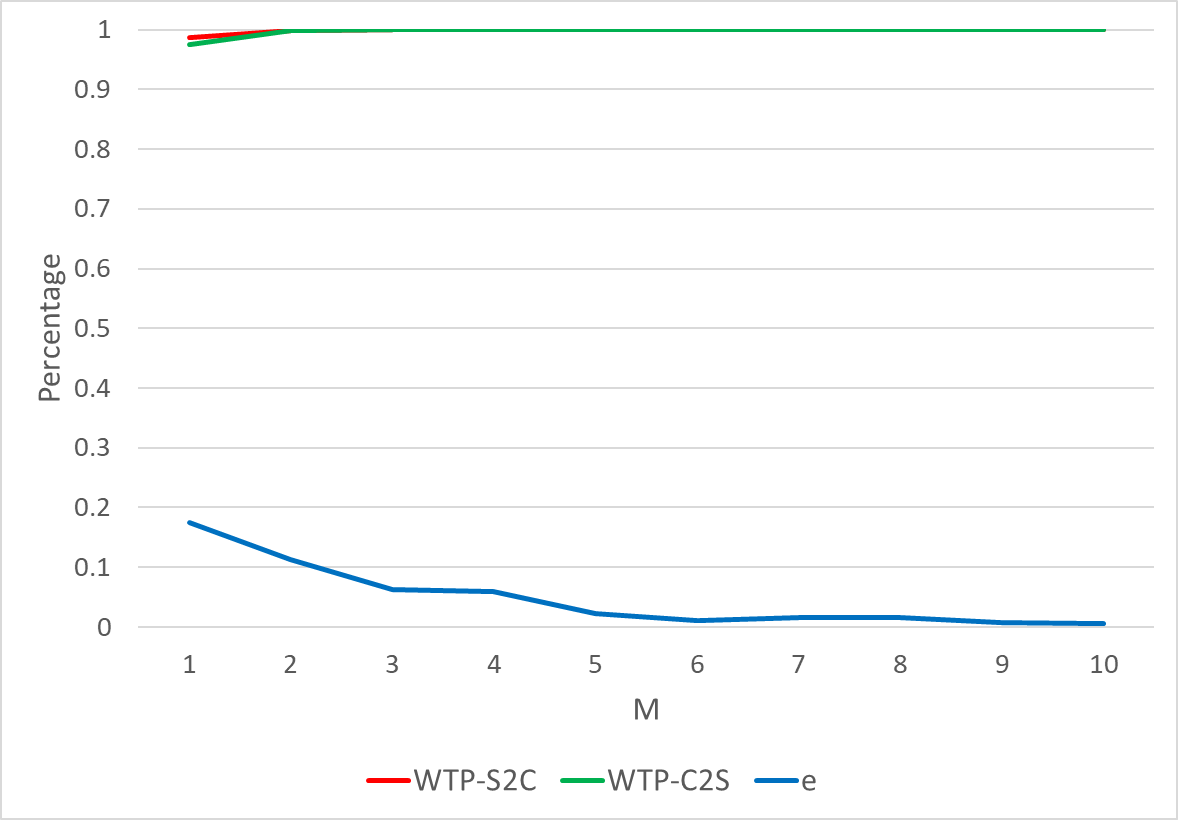}
	\caption{The measures of accuracy WTP-C2S, WTP-S2C and $e$ plotted against different values of $M$ for PMIT-S0L+PL.}
	\label{fig:pmits0lresult4}
\end{figure}

The main research goal of this paper is investigating the effect of inferring an S0L-system when using various numbers of sequences of strings as an input. Figure \ref{fig:pmits0lresult4} shows how the probability error, WTP-C2S and WTP-S2C change as $M$ increases. This raw data is shown in Table \ref{table:results5a}, which also shows the mean time to solve, the maximum successor difference, and successor difference rate. SR is not shown as it was always $100\%$, for each value of $M$ with PMIT-S0L+PL and dataset $DS_{vM}$. It can been seen that from $M \geq 3$, PMIT-S0L+PL infers the original system for all test cases, as the WTP values are all $1.000$. With respect to error in the probability distribution, this becomes subjectively reasonable at $M=3$ with $6.3\%$ average error. Error seems to be close to minimal at $M=6$ (at approximately $1\%$), although it continues to decline at a small rate as $M$ increases. Increasing $M$ has a negligible effect on MTTS since $\rho_{2}$ to $\rho_{M}$ are only scanned when an S0L-system has been found to be compatible with $\rho_{1}$, and this is similar to the scanning process which takes sub-millisecond time in all tests cases. So, adding additional sequences of strings is generally beneficial in practice. 

With respect to differences between the candidate system and the original system, it can be seen that when $M=1$ there can be up to two successor differences. These non-matching productions tend to be for low probability successors as can be seen from the values for WTP. When $M=2$, there was a single instance where PMIT-S0L+PL did not exactly match the original system. Again, the difference was a very low probability successor. In examining this experiment more closely, it was found that the successor in the original system had only been selected once and the symbols lined up in such a fashion that the candidate system found by PMIT-S0L+PL was reasonable given the data. Still, overall, in practice it would be recommended to have at least $3$ sequences of strings to infer an S0L-system that closely models an underlying process reliably.

	\begin{table}
		\centering
		\begin{tabular}{|c|c|c|c|c|c|c|}
			\hline
			$M$  & WTP-S2C & WTP-C2S & $e$     & \textit{diffmax} & \textit{diffrate}  & MTTS \\ \hline
			1    & $0.987$     & $0.975$    & $0.174$ & $2$          & $20\%$         & $03\text{:}48.228$ \\ \hline
			2    & $0.998$     & $0.999$    & $0.113$ & $1$          & $5\%$          & $01\text{:}35.192$ \\ \hline
			3    & $1.000$     & $1.000$    & $0.063$ & $0$          & $0\%$          & $01\text{:}30.821$ \\ \hline
			4    & $1.000$     & $1.000$    & $0.060$ & $0$          & $0\%$          & $03\text{:}55.821$ \\ \hline
			5    & $1.000$     & $1.000$    & $0.023$ & $0$          & $0\%$          & $02\text{:}30.821$ \\ \hline
			6    & $1.000$     & $1.000$    & $0.010$ & $0$          & $0\%$          & $04\text{:}00.821$ \\ \hline
			7    & $1.000$     & $1.000$    & $0.016$ & $0$          & $0\%$          & $02\text{:}16.131$ \\ \hline
			8    & $1.000$     & $1.000$    & $0.016$ & $0$          & $0\%$          & $02\text{:}16.131$ \\ \hline
			9    & $1.000$     & $1.000$    & $0.008$ & $0$          & $0\%$          & $02\text{:}46.496$ \\ \hline
			10   & $1.000$     & $1.000$    & $0.006$ & $0$          & $0\%$          & $03\text{:}00.065$ \\ \hline
		\end{tabular}
		\caption{Performance metrics for PMIT-S0L+PL with $M$ sequences of strings from $1$ to $10$, $N=S$, and the dataset $DS_{vM}$. $\rm SR = 100\%$ for all experiments and is therefore not in the table.}
		\label{table:results5a}
	\end{table}\textbf{}

	With respect to inferring the S0L-system found by Nishida \cite{nishida_k0l} for modeling Japanese Cypress, PMIT-S0L+PL was not able to infer it in a practical amount of time. The first experiment was to set $N=42$, which is much greater than the approximate maximum of $9$ successors in the other experiments. After several hours, this was terminated and it was estimated that PMIT-S0L+PL would take at least $10^9$ hours to complete using exhaustive search in a sequential fashion.

	\section{Conclusions and Future Directions}\label{section:conclusions}
	
	This paper presents an investigation into inferring stochastic context-free L-systems (S0L-systems) when using different numbers of sequences of strings with the Plant Model Inference Tool for S0L-systems (PMIT-S0L). PMIT-S0L is a generalized algorithm for inferring S0L-system and requires no \textit{a priori} scientific knowledge when compared to existing approaches for inferring S0L-systems algorithmically (e.g. \cite{danks_s0lproteinfolding,beauty,rongier_sedimentary}). Being generalized means that PMIT-S0L may be used for any problem as opposed to requiring a specific algorithm for each individual problem in a specific research domain. PMIT-S0L opens up the possibility of inferring S0L-systems in research domains where none have been found to date.
	
	PMIT-S0L is primarily evaluated on procedurally generated S0L-systems due a shortage of specific systems published in the literature. An analysis was done of existing L-systems to create realistic procedural generation rules. Three data sets, for a total of $960$ L-systems (a total of $3,000$ experiments were conducted across these systems), were generated to evaluate different aspects of PMIT-S0L.
	
	PMIT-S0L has two different modes of operation controlled by a \textit{prefix limitation} Boolean parameter (denoted as PMIT-S0L+PL and PMIT-S0L-PL). The \textit{prefix limitation} parameter controls if it searches all L-systems, or only those without multiple successors of the same letter, where one is a prefix of the other. The results show PMIT-S0L-PL is quite a bit faster than PMIT-S0L+PL. For an S0L-system with $9$ successors, PMIT-S0L-PL will succeed on average in about $20$ minutes, while PMIT-S0L+PL takes several hours. However, PMIT-S0L+PL is still practical considering the effort required to infer an S0L-system by hand.
	
	PMIT-S0L-PL was found be an extremely accurate tool for inferring a compatible S0L-system that is at least as probable as the original system. This paper shows that PMIT-S0L+PL was always able to find the original system when at least three sequences of strings were used as inputs. Additionally, the evaluation showed that when six sequences were used, the error in the solution's probability distribution compared to the original system becomes low (about $1\%$ or less), and there is not much improvement for adding additional sequences of strings. Therefore, it is recommended that in practice at least $3$ sequences of strings should be used to infer S0L-systems, but $6$ or more is ideal to minimize the error in probabilities. Finally, there is essentially no penalty to execution time for adding additional sequences of strings, so there is little reason to avoid using all of the string sequences that are available.
	
	PMIT-S0L (in either mode) can be used with an exhaustive search, which is not an efficient searching algorithm. However, with normal branch-and-bound pruning, it could infer all L-systems in a test suite of $420$ L-systems with at most $9$ productions in about $8$ hours. Genetic algorithm was also evaluated, and was much faster; however, it was less accurate.
	
	There are several directions that can be taken with this research in the future. Perhaps most importantly comes from applying this research to the practical problem of inferring L-systems from segmented images. De La Higuera \cite{de2005bibliographical} argues that to be realistic, L-system inference algorithms should handle errors in the strings; e.g., insertions or deletions of symbols. These errors can be, at least initially, considered stochastic productions; i.e., if $A$ has the production $A \rightarrow xyz$, but periodically is subject to a deletion error such that $A \rightarrow xz$, then this can be thought of as a stochastic production. With a set of stochastic productions inferred, it is then a matter of detecting and fixing any errors.
	
	Additionally, parallelism will be used to speed up PMIT-S0L; however, it still would be useful to investigate more efficient searching techniques to allow PMIT-S0L to infer S0L-systems with larger number of productions at practical speeds. Additionally, the main issue with using PMIT-S0L in a practical fashion is the need to select a reasonable value for the size of the vector used for searching (which roughly corresponds to the number of productions). An algorithm that does not require this parameter would be ideal, or alternatively finding a good way to compute or estimate the vector size.
	
	\section{Acknowledgments}
	
	This research was undertaken thanks in part to funding from the Canada First Research Excellence Fund, National Science Engineering Research Council grant \#2016-06172, and the Alexander Graham Bell scholarship for Jason Bernard. Additionally, the authors would like to thank Dr.\ Farhad Maleki for his assistance in acquiring the images for this paper.
	
	\bibliographystyle{elsarticle-num}
	\bibliography{pmit_sol}{}
	
\end{document}